\documentclass{article}

\usepackage[final]{nips_2018}

\usepackage{subfig}

\usepackage{float}

\usepackage[utf8]{inputenc} 
\usepackage[T1]{fontenc}    
\usepackage{hyperref}       
\usepackage{url}            
\usepackage{booktabs}       
\usepackage{amsfonts}       
\usepackage{nicefrac}       
\usepackage{microtype}      

\usepackage{graphicx}

\usepackage{multirow}

\usepackage{amsmath}

\usepackage{amssymb}

\usepackage[table]{xcolor}
\usepackage{collcell}
\usepackage{hhline}
\usepackage{pgf}

\def\colorModel{rgb} 

\newcommand\ColCell[1]{
  \pgfmathparse{#1<50?1:0}  
    \ifnum\pgfmathresult=0\relax\color{white}\fi

\pgfmathsetmacro\compA{100(1-#1/100)}      
\pgfmathsetmacro\compB{100(1-#1/100)} 
\pgfmathsetmacro\compC{100(1-#1/100)}      

  \edef\x{\noexpand\centering\noexpand\cellcolor[\colorModel]{\compA,\compB,\compC}}\x #1
  }
\newcolumntype{E}{>{\centering\arraybackslash\collectcell\ColCell}m{0.65cm}<{\endcollectcell}}  

\begin{document}

\title{Real-time Distracted Driver Posture Classification}
\author{Yehya Abouelnaga\\
Department of Informatics\\
Technical University of Munich\\
\texttt{yehya.abouelnaga@tum.de}
\And
Hesham M. Eraqi\\
Department of Computer Science and Engineering\\
The American University in Cairo\\
\texttt{heraqi@aucegypt.edu}
\And
Mohamed N. Moustafa\\
Department of Computer Science and Engineering\\
The American University in Cairo\\
\texttt{m.moustafa@aucegypt.edu}
}

\maketitle

\begin{abstract}
In this paper, we present a new dataset for ``distracted driver'' posture estimation.
In addition, we propose a novel system that achieves 95.98\% driving posture estimation classification accuracy.
The system consists of a genetically-weighted ensemble of Convolutional Neural Networks (CNNs). We show that a weighted ensemble of classifiers using a genetic algorithm yields in better classification confidence.
We also study the effect of different visual elements (i.e.\ hands and face) in distraction detection and classification by means of face and hand localizations.
Finally, we present a thinned version of our ensemble that could achieve a 94.29\% classification accuracy and operate in a realtime environment.
\end{abstract}


\section{Introduction}

The number of road accidents due to distracted driving is steadily increasing.
According to the National Highway Traffic Safety Administration (NHTSA), in 2015, 3,477 people were killed, and 391,000 were injured in motor vehicle crashes involving distracted drivers \citet{USDepartmentofTrans2017}.
The major cause of these accidents was the use of mobile phones.
The NHTSA defines distracted driving as ``any activity that diverts attention from driving'', including: a) Talking or Texting on one's phone, b) eating and drinking, c) talking to passengers, d) fiddling with the stereo, entertainment, or navigation system \citet{USDepartmentofTrans2017}.
The Center for Disease Control and Prevention (CDC) provides a broader definition of distracted driving by taking into account visual (i.e.\ taking one's eyes off the road), manual (i.e.\ taking one's hands off the driving wheel) and cognitive (i.e.\ taking one's mind off driving) causes \citet{Services2016}.
We believe that the detection of distracted driver's postures is key to further preventive measures.
Distracted driver detection is also important for autonomous vehicles; Latest commercial  self-driving cars still require drivers to pay attention and be ready to take back control of the wheel \citet{eriksson2017takeover}.

We present a realtime distracted driver pose estimation system using a weighted ensemble of convolutional neural networks and a challenging distracted driver's dataset on which we evaluate our proposed solution.

\section{Literature Review}

The work in the distracted driver detection field over the past seven years could be clustered into four groups: multiple independent cell phone detection publications, Laboratory of Intelligent and Safe Automobiles in University of California San Diego (UCSD) datasets and publications, Southeast University Distracted Driver dataset and affiliated publications, and recently, StateFarm's Distracted Driver Kaggle competition.

\subsection{Cell Phone Usage Detection}

\citet{Berri2014} presents an SVM-based model that detects the use of mobile phone while driving (i.e.\ distracted driving).
Their dataset consists of frontal image view of a driver's face.
They also make pre-made assumptions about hand and face locations in the picture.
\citet{craye2015driver} uses AdaBoost classifier and Hidden Markov Models to classify a Kinect's RGB-D data.
Their solution depends on data produced by indoor data.
They sit on a chair and a mimmic a certain distraction (i.e.\ talking on the phone).
This setup misses two essential points: the lighting conditions and the distance between a Kinect and the driver.
In real-life applications, a driver is exposed to a variety of lighting conditions (i.e.\ sunlight and shadow).
\citet{HoangNganLe2016} devised a Faster-RCNN model to detect driver's cell-phone usage and ``hands on the wheel''.
Their model is mainly geared towards face/hand segmentation.
They train their Faster-RCNN on the dataset proposed in \citet{Das2015} (that we also use in this paper).
Their proposed solution runs at a 0.06, and 0.09 frames per second for cell-phone usage, and ``hands on the wheel'' detection.

\subsection{UCSD's Laboratory of Intelligent and Safe Automobiles Work}


In \citet{ohn2013driver}, the authors present a fusion of classifiers where they segment the image to three regions: wheel, gear, and instrument panel (i.e.\ radio).
They develop a classifier for each segment in which they detect existence of hands in those areas.
The information from these scenes are passed to an ``activity classifier'' that detects the actual activity (i.e.\ adjusting the radio, operating the gear).
\citet{Ohn-bar2013} presents a region-based classification approach.
It detects hands presence in certain pre-defined regions in an image.
A model is learned for each region separately.
All regions are later joined using a second-stage classifier.
\citet{Ohn-bar2013a} collects a new RGBD dataset in which they observe the driving wheel and a driver's hand activity.
The frames are divided into 5 labelled regions with classes: One hand, no hands, two hands, two hands + cell, two hands + map, and two hands + bottle.

\subsection{Southeast University Distracted Driver Dataset}


\citet{zhao2012recognitionRF} designed a more inclusive distracted driving dataset with a side view of the driver and more activities: grasping the steering wheel, operating the shift lever, eating a cake and talking on a cellular phone.
In their paper, they introduced a contourlet transform for feature extraction, and then, evaluated the performance of different classifiers: Random Forests (RF), $k$-nearest neighbors classifier (KNN), and Multi-Layer Perceptron (MLP) classifier.
The random forests achieved the highest classification accuracy of 90.5\%.
\citet{zhao2012recognitionMLP} showed that using a multiwavelet transform improves the accuracy of multilayer perceptron classifier to 90.61\% (previously 37.06\%).
\citet{zhao2013recognition} improves the Multilayer Perceptron (MLP) classifier using combined features of Pyramid Histogram of Oriented Gradients (PHOG) and spatial scale feature extractors.
Their MLP achieves a 94.75\% classification accuracy.
\citet{Yan2016} introduces a R*CNN that trains on manually labelled pre-defined regions (i.e.\ driver, shift lever).
Their convolutional nerual net achieves a 97.76\%.
It is worth noting that all previous publications tested their accuracies against four classes.
This publication tested against six classes.
\citet{Yan2016DrivingDistraction} presents a convolutional neural network solution that achieves a 99.78\% classification accuracy.
They train their network in a 2-step process. First, they use pre-trained sparse filters as the parameters of the first convolutional layer.
Second, they fine-tune the network on the actuall dataset.
Their accuracy is measured against the 4-classes of the Southeast dataset: wheel (safe driving), eating/smoking, operating the shift lever, and talking on the phone.

\subsection{StateFarm's Dataset}

StateFarm's Distracted Driver Detection competition on Kaggle was the first publicly available dataset for posture classification.
In the competition, StateFarm defined ten postures to be detected: safe driving, texting using right hand, talking on the phone using right hand, texting using left hand, talking on the phone using left hand, operating the radio, drinking, reaching behind, hair and makeup, and talking to passenger.
Our work, in this paper, is mainly inspired by StateFarm's Distracted Driver's competition.
While the usage of StateFarm's dataset is limited to the purposes of the competition \citet{Sultan2016}, we designed a similar dataset that follows the same postures.


\section{Dataset Design}

\begin{figure*}[th]
  \centering
  \includegraphics[width=\textwidth]{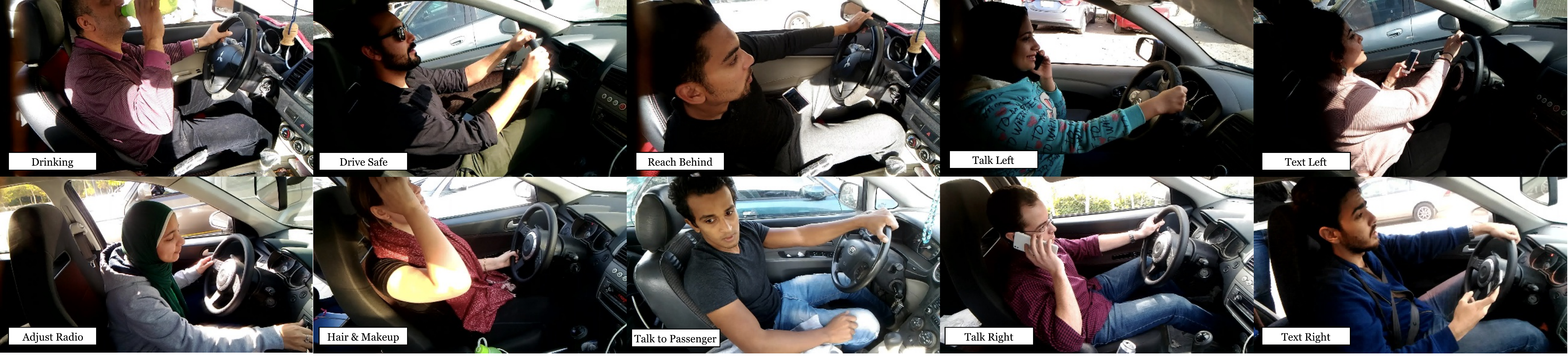}
  \caption{Examples of the American University in Cairo (AUC) Distracted Driver's Dataset. In a column-level order, postures are: drinking, adjusting the radio, driving in a safe posture, fiddling with hair or makeup, reaching behind, talking to passengers, talk on cell phone using left hand, talk on cell phone using right hand, texting using left hand, and texting using right hand.}
\end{figure*}

Creating a new dataset (``AUC Distracted Driver'' dataset) was essential to the completion of this work.
The available alternatives to our dataset are: StateFarm and Southeast University (SEU) datasets. StateFarm's dataset is to be used for their Kaggle past competition purpose only (as per their regulations) \citet{Sultan2016}. As per our multiple attempts to obtain it, we knew that the authors of Southeast University (SEU) dataset do not make it publicly available. Also, their dataset consists of only four postures. All the papers (\citet{Yan2016,Yan2016DrivingDistraction,Yan2014,zhao2013recognition,zhao2012recognitionMLP,Zhao2011,zhao2012recognitionRF}) that benchmarked against the dataset are affiliated with the either Southeast University, Xi’an Jiaotong-Liverpool University, or Liverpool University, and they have at least one shared author.


The dataset was collected using an ASUS ZenPhone (Model Z00UD) rear camera.
The input was collected in a video format, and then, cut into individual images, $1080 \times 1920$ each.
The phone was fixed using an arm strap to the car roof handle on top of the passenger's seat.
In our use case, this setup proved to be very flexible as we needed to collect data in different vehicles.
In order to label the collected videos, we designed a simple multi-platform action annotation tool.
The annotation tool is open-source and publicly available at \citet{Abouelnaga2017}.



We had 31 participants from 7 different countries: Egypt (24), Germany (2), USA (1), Canada (1), Uganda (1), Palestine (1), and Morocco (1).
Out of all participants, 22 were males and 9 were females.
Videos were shot in 4 different cars: Proton Gen2 (26), Mitsubishi Lancer (2), Nissan Sunny (2), and KIA Carens (1).


\section{Proposed Method}

\begin{figure*}
  \centering
  \includegraphics[width=\textwidth]{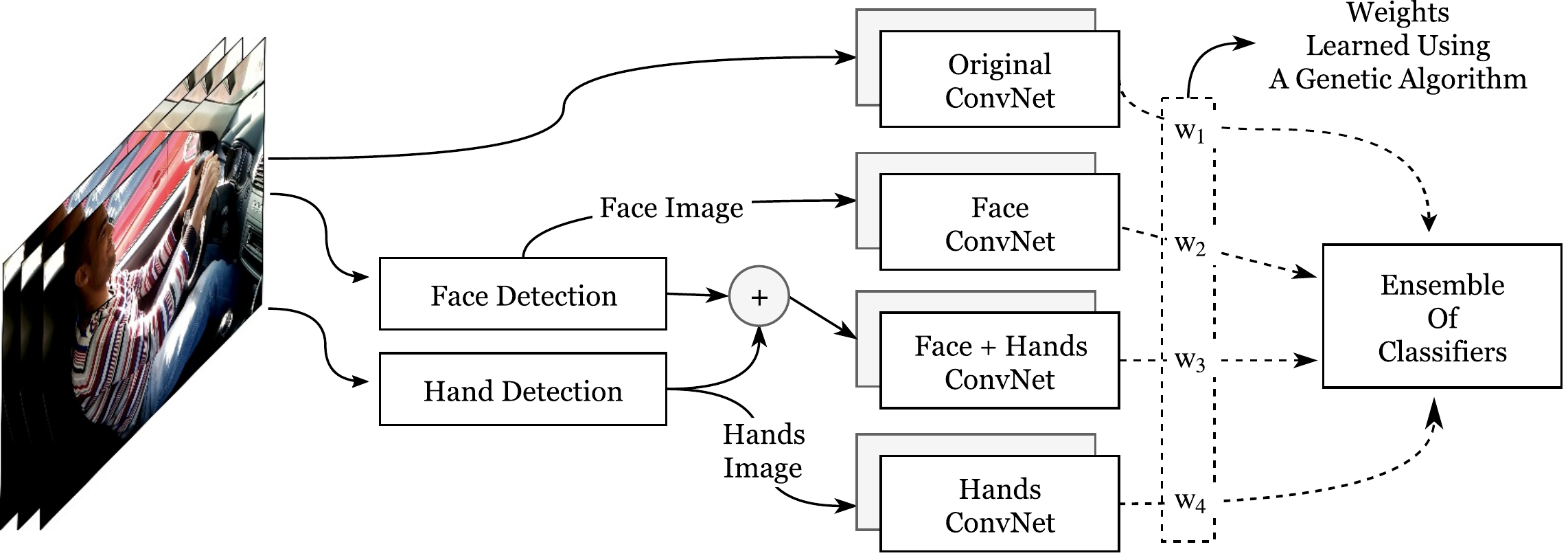}
  \caption{An overview of our proposed solution.
      A face detector and a hand detector are run against each frame.
      For each output image (i.e. Face and Hands), an AlexNet and an InceptionV3 networks are trained (i.e.\ resulting in 8 neural networks: 4 AlexNet and 4 InceptionV3).
      The overall class distribution is determined by the weighted sum of all softmax layers.
      The weights are learned using a genetic algorithm.
  }
\end{figure*}

Our proposed solution consists of a genetically-weighted ensemble of convolutional neural networks.
The convolutional neural networks train on raw images, face images, hands images, and ``face+hands'' images.
We train an AlexNet \citet{krizhevsky2012imagenet} and an InceptionV3 \citet{szegedy2016rethinking} on those four images sources.
In the InceptionV3 network, we fine-tune a pre-trained ImageNet model (i.e.\ transfer learning).
Then, we evaluate a weighted sum of all networks' outputs yielding the final class distribution.
The weights are evaluated using a genetic algorithm.

\subsection{Face \& Hands Detection}
We trained the model presented in \citet{LiHaoxiangandLinZheandShenXiaohuiandBrandtJonathanandHua2015} on the Annotated Facial Landmarks in the Wild (AFLW) face dataset \citet{tugraz:icg:lrs:koestinger11b}.
The trained model achieved decent results.
However, it was sensitive to distance from the camera (i.e.\ faces that were close to the camera were not easily detected).
We found that the pre-trained model (presented in \citet{Farfade2015}) produced better results on our dataset.
Given that we did not have any hand labelled face bounding boxes, we couldn't formally compare the two models.
However, when randomly selecting images from different classes, we found that \citet{Farfade2015} was closer to what we expected.

As for hands detection, we used the pre-trained model presented in \citet{Bambach_2015_ICCV} with slight modifications.
Their trained model was a binary class AlexNet that classifies hands/non-hands for different proposal windows.
We transferred the weights of the fully connected layers (i.e.\ fc6, fc7 and fc8) into convolutional layers such that each neuron in the fully connected layer was transferred into a depth layer with a 1-pixel kernel size. Except the first fully connected layer.
Also, this architecture accepts variant size inputs and produces variant-size outputs.
The last convolutional layer has a depth of 2 (i.e.\ the binary classes) where $\text{Conv8}_{x,y,0} + \text{Conv8}_{x,y,1} = 1
$ for all $x$ and $y$; such that $0 \le x < W$, $0 \le y < H$ and $W$ and $H$ are the output's width and height, respectively.

\subsection{Convolutional Neural Network}

For distracted driver posture classification, we trained two classes of neural networks: AlexNet and InceptionV3.
Each network is trained on 4 different image sources (i.e.\ raw, face, hands and face+hands images) yielding in 4 models per net and a total of 8 models.

We trained our AlexNet models from scratch. We didn't use a pre-trained model.
For InceptionV3, we performed a transfer learning.
We fine-tuned a pre-trained model on the distraction postures.
We removed the ``logits'' layer, and replaced it with a 10-neuron fully connected layer (i.e.\ corresponding to 10 driving postures).

We used a gradient descent optimizer with an initial learning rate of $10^{-2}$.
The learning rate decays linearly in each epoch with a step of $(10^{-2} - 10^{-4}) / Epochs$.
We trained the networks for 30 epochs. In each, we divide the training dataset into mini-batches of 50 images each.


\subsection{Weighted Ensemble of Classifiers using Genetic Algorithm}
Each classifier produces a class probability vector (i.e.\ output of ``softmax'' layer), $C_1 \,\ldots\,\, C_N$, such that $C_i$ has 10 probabilities (i.e.\ 10 classes) and $N$ is the number of classifiers ($N = 8$ in our situation).
In a majority voting system, we assume that all experts (i.e.\ classifiers) can equally contribute to a better decision by taking the unweighted sum of all classifier outputs.
\[
C_{\text{Majority}} = \frac{1}{N} \sum_{i}^{N} C_i, \,\,\,\,\,\,
C_{\text{Weighted}} = \frac{1}{\sum_{i}^{N} w_i} \sum_{i}^{N} w_i \cdot C_i
\]
However, that is not usually a valid assumption.
In a weighted voting system, we assume that classifiers do not contribute equally to the ensemble and that some classifiers might yield higher accuracy than others.
Therefore, we need to estimate the weights of each classifier's contribution to the ensemble.
\citet{rokach2010ensemble} presents a variety of methods to estimate the weights.
We opted to use a genetic algorithm (i.e.\ a search-based method).

Our chromosome consists of $N$ genes that correspond to the weights $w_1 \,\, \ldots \,\, w_N$.
Our fitness function evaluates the Negative Log Likelihood (NLL) loss over a 50\% random sample of the population.
This helps prevent overfitting.
Our population consists of 50 individual.
In each iteration, we retain the top 20\% of the population and use them as parents.
Then, we randomly select 10\% of the remaining 80\% of the population as parents.
In other words, we have 30\% of the population as parents.
Now, we randomly mutate 5\% of the selected parents.
Finally, we cross-over random pairs of the parents to produce children until we have a full population (i.e.\ with 50 individuals).
We ran the above procedure for only 5 iterations in order to avoid over-fitting.
We selected the chromosome with the highest fitness score (test against all data points-- not 50\%).

\section{Experiments}


\def\arraystretch{1.5} 
\begin{table}[t]
\centering
\arrayrulecolor{black} 
\caption{Distracted Driver Posture Classification Results}
\label{classification-results}
\begin{tabular}{|c|c|c|c|}
\hline
Model                            & Source               & Loss (NLL)            & Accuracy (\%)  \\ \hline
\multirow{5}{*}{AlexNet}         & \textbf{Original}    & \textbf{0.3909} & \textbf{93.65} \\ \cline{2-4}
                                 & Face                 & 1.0516          & 84.28          \\ \cline{2-4}
                                 & Hands                & 0.6186          & 89.52          \\ \cline{2-4}
                                 & Face + Hands         & 0.8298          & 86.68          \\ \hline
\multirow{5}{*}{InceptionV3}     & \textbf{Original}    & \textbf{0.2654} & \textbf{95.17} \\ \cline{2-4}
                                 & Face                 & 0.6096          & 88.82          \\ \cline{2-4}
                                 & Hands                & 0.4546          & 91.62          \\ \cline{2-4}
                                 & Face + Hands         & 0.4495          & 90.88          \\ \hline
\multicolumn{2}{|c|}{Realtime System} & 0.2727 & 94.29 \\ \hline
\multicolumn{2}{|c|}{\textbf{Majority Voting Ensemble}} & \textbf{0.1661} & \textbf{95.77} \\ \hline
\multicolumn{2}{|c|}{\textbf{GA-Weighted Ensemble}}     & \textbf{0.1575} & \textbf{95.98} \\ \hline
\end{tabular}
\end{table}
  
%

We divided our dataset into 75\% training and 25\% held out test data.
Then, we ran the face and hand detectors on the entire dataset.
We tested all of the networks against our test dataset and obtained the results in Table~\ref{classification-results}.
We notice that both AlexNet and InceptionV3 achieve best accuracies when trained on the original images.
Hands seem to have more weight in posture recognition than the face.
``Face + Hands'' images produce slightly lower accuracy than the hands images, yet, still higher than the face images.
That happens due to face/hand detector failures.
For example, if a hand is not found, we pass a face image to a ``face + hands'' classifier.
This doesn't happen in individual cases of hand-only or face-only classifier because if the hand/face detection fails, we pass the original image to the hand/face classifier as a fallback mechanism.
With better hand/face detectors, the ``face+hands'' networks are expected to produce higher accuracies than the ``hands'' networks.
An ensemble of two AlexNet models produce a satisfactory classification accuracy (i.e.\ 94.29\%).
Meanwhile, it still maintains a realtime performance on a CPU-based system.

We trained and tested our models using an EVGA GeForce GTX TITAN X 12GB GPU, Intel(R) Core(TM) i7-5960X CPU @ 3.00GHz, and a 48 GM RAM\@.
On average, AlexNet processed 182 frames per second using a GPU and 52 frames per second using a CPU\@.
InceptionV3 processes 72 frames per second using a GPU and 5.5 frames per second using a CPU.


\begin{table}[t]
\def\arraystretch{1.5} 
\small
\caption{Confusion Matrix of Genetically Weighted Ensemble of Classifiers}
\label{fig:ga-confusion-matrix}
\centering
\newcommand\items{10}   
\arrayrulecolor{white} 
\noindent\begin{tabular}{cc*{\items}{|E}|}
\centering
\tabularnewline
\multicolumn{1}{c}{} &\multicolumn{1}{c}{} &\multicolumn{\items}{c}{Predicted} \\ \hhline{~*\items{|-}|}
\multicolumn{1}{c}{} & 
\multicolumn{1}{c}{} & 
\multicolumn{1}{c}{C0} & 
\multicolumn{1}{c}{C1} & 
\multicolumn{1}{c}{C2} & 
\multicolumn{1}{c}{C3} & 
\multicolumn{1}{c}{C4} & 
\multicolumn{1}{c}{C5} & 
\multicolumn{1}{c}{C6} & 
\multicolumn{1}{c}{C7} & 
\multicolumn{1}{c}{C8} & 
\multicolumn{1}{c}{C9} \\ \hhline{~*\items{|-}|}
\multirow{\items}{*}{\rotatebox{90}{Actual}} 
&C0     &95.34  &0      &0.33   &0.65   &0.11   &0.43   &0.43   &0.87   &0.11   &1.74   \\ \hhline{~*\items{|-}|}
&C1     &0.31   &96.63  &1.23   &0.31   &0.92   &0      &0.31   &0      &0.31   &0      \\ \hhline{~*\items{|-}|}
&C2     &0.29   &3.23   &96.48  &0      &0      &0      &0      &0      &0      &0      \\ \hhline{~*\items{|-}|}
&C3     &2.02   &0.61   &0      &96.15  &0.81   &0      &0.20   &0      &0      &0.20   \\ \hhline{~*\items{|-}|}
&C4     &0      &0.33   &0      &4.90   &94.77  &0      &0      &0      &0      &0      \\ \hhline{~*\items{|-}|}
&C5     &4.26   &0      &0      &0.33   &0      &95.08  &0      &0      &0      &0.33   \\ \hhline{~*\items{|-}|}
&C6     &0.74   &0      &0      &0.25   &0      &0.74   &98.01  &0.25   &0      &0      \\ \hhline{~*\items{|-}|}
&C7     &3.65   &0      &0      &0      &0      &0      &0      &95.35  &0      &1.00   \\ \hhline{~*\items{|-}|}
&C8     &3.79   &0      &0      &0      &0      &0      &1.38   &0.34   &92.76  &1.72   \\ \hhline{~*\items{|-}|}
&C9     &1.40   &0      &0      &0      &0      &0      &0.47   &0.31   &0.16   &97.67  \\ \hhline{~*\items{|-}|}
\end{tabular}
\end{table}

\subsection{Analysis}
In Table \ref{fig:ga-confusion-matrix}, we notice that the most confusing posture is the ``safe driving''.
This is due to the lack of temporal context in static images.
In a static image, a driver would appear in a ``safe driving'' posture.
However, contextually, he/she was distracted by doing some other activity.
``Text Left'' is mostly confused for ``Talk Left'' and vice versa.
Same applies to ``Text Right'' and ``Talk Right''.
``Adjust Radio'' is mainly confused for a ``safe driving'' posture.
That is due to lack of the previously mentioned temporal context.
Apart from safe driving, ``Hair \& Makeup'' is confused for talking to passenger.
That is because, in most cases, when drivers did their hair/makeup on the left side of their face, they needed to tilt their face slightly right (while looking at the frontal mirror).
Thus, the network thought the person was talking to passenger.
``Reach Behind'' was confused for both talking to passenger and drinking.
That makes sense as people tend to naturally look towards the camera while reaching behind.
As for the drinking confusion, it is due to right-arm movement from the steering wheel to the back seat.
A still image in the middle of that move could be easily mistaken for a drinking posture.
``Drink'' and ``Talk to Passenger'' postures were not easily confused with other postures as 98\% and 97.67\% of their images were correctly classified.


\section{Conclusion}
Distracted driving is a major problem leading to a striking number of accidents worldwide.
In addition to regulatory measures to tackle such problems, we believe that smart vehicles would indeed contribute to a safer driving experience.
In this paper, we presented a robust vision-based system that recognizes distracted driving postures.
We collected a challenging distracted driver dataset that we used to develop and test our system.
Our best model utilizes a genetically weighted ensemble of convolutional neural networks to achieve a 95.98\% accuracy.
We also showed that a simpler model (only using AlexNet) could operate in realtime and still maintain a satisfactory classification accuracy.
Face and hands detection is proved to improve classification accuracy in our ensemble.
However, in a realtime setting, their performance overhead is much higher than their contribution.

In a future work, we need to devise a better face and hands detector.
We would need to manually label hand and face proposals and use them to train an object detector (i.e. SSD) to improve faces and hands localization.
In order to overcome the ``safe driving'' posture confusion with other classes, we would need to incorporate temporality in our decision.
We shall test the performance of a Recurrent Neural Network (RNN) against sequential stream of frames.
We envision a performance improvement due to temporal features.


\medskip

\bibliographystyle{unsrtnat}
\bibliography{library}

\end{document}